\documentclass[
    sigconf = true,     
    review = false,      
    screen = true,      
    anonymous = false,   
]
{acmart}

\usepackage{amsmath,amssymb,amsthm}
\usepackage{xspace}
\usepackage{epsfig}

\usepackage{algorithm}
\usepackage[noend]{algorithmic}
 \usepackage{color}
\usepackage{url}
\usepackage{booktabs,multirow}
\usepackage{tikz}
\usepackage{verbatim}
\usetikzlibrary{arrows,shapes}
\usepackage{tabularx}

\usepackage{graphicx}
\usepackage[normalem]{ulem}
\useunder{\uline}{\ul}{}

\renewcommand{\epsilon}{\varepsilon}

\begin{document}

\copyrightyear{2020}
\acmYear{2020}
\setcopyright{none}
\acmConference[GECCO '20]{Genetic and Evolutionary Computation Conference}{July 8--12, 2020}{Cancún, Mexico}
\acmBooktitle{Genetic and Evolutionary Computation Conference (GECCO '20), July 8--12, 2020, Cancún, Mexico}
\acmPrice{15.00}
\acmDOI{10.1145/3377930.3390183}
\acmISBN{978-1-4503-7128-5/20/07}





\author{Anja Jankovic}
\affiliation{%
  \institution{Sorbonne Universit\'e, CNRS, LIP6}
  \city{Paris, France}
}

\author{Carola Doerr}
\affiliation{%
  \institution{Sorbonne Universit\'e, CNRS, LIP6}
  \city{Paris, France}
}

\title[Performance Regression for the Modular CMA-ES]{Landscape-Aware Fixed-Budget Performance Regression\\ and Algorithm Selection for Modular CMA-ES Variants}

\begin{abstract}
Automated algorithm selection promises to support the user in the decisive task of selecting a most suitable algorithm for a given problem. A common component of these machine-trained techniques are regression models which predict the performance of a given algorithm on a previously unseen problem instance. In the context of numerical black-box optimization, such regression models typically build on exploratory landscape analysis (ELA), which quantifies several characteristics of the problem. These measures can be used to train a supervised performance regression model. 

First steps towards ELA-based performance regression have been made in the context of a fixed-target setting. In many applications, however, the user needs to select an algorithm that performs best within a given budget of function evaluations. Adopting this fixed-budget setting, we demonstrate that it is possible to achieve high-quality performance predictions with off-the-shelf supervised learning approaches, by suitably combining two differently trained regression models. 
We test this approach on a very challenging problem: algorithm selection on a portfolio of very similar algorithms, which we choose from the family of modular CMA-ES algorithms.  
\end{abstract}


\maketitle

\sloppy{

\section{Introduction}

In the vast realm of optimization problems, we often encounter real-world challenges that are too complex to be analytically modeled, but that we nevertheless need to find an optimal solution for.  Under these circumstances, one typically resorts to using \emph{black-box} optimization techniques, which guide the search towards an estimated optimal solution, iteration by iteration, using only pairs of potential candidate solutions $x$ and their corresponding fitness values $f(x)$ in each step.
Many different iterative optimization heuristics have been designed in the last decades, and continue to be developed every day, mostly because these heuristics show (very) different performances on different problems. We know today, as the \emph{no free lunch theorem}~\cite{NFL} shows, that the quest for a single-best optimization algorithm, the one that would optimally (or quasi-optimally) solve any kind of problem in the most time- and resource-efficient way possible, is futile. 
Consequently, every time one is faced with a previously unseen 
optimization problem, the most appropriate algorithm for that specific scenario must first be selected.  This so-called \emph{algorithm selection problem} or \emph{ASP} was formalized a few decades ago~\cite{rice1976asp}, and the community has during that time focused predominantly on algorithm selection in the field of discrete optimization~\cite{xhhlb2011satzilla}. In the last decade, however, substantial progress has been made also for numerical optimization problems~\cite{kerschke2018survey}.  

A special case of the ASP is \emph{per instance algorithm selection}, or PIAS for short, where a most suitable algorithm for a given problem instance is to be chosen from a discrete (or at least countable) algorithm portfolio. A key component in designing PIAS are mechanisms that are able to characterize or to identify the problem instance at hand.  In the continuous domain, this typically means that we need to recognize how the fitness landscape of a problem looks like.  It is important to mention that high-level, intuitive problem landscape properties (e.g., degree of multimodality, separability, number of plateaus) all require prior expert knowledge and cannot be automatically computed.  With the goal of describing the problem by means of numerical values, we turn to \emph{Exploratory Landscape Analysis}, which was introduced in~\cite{mersmann2011exploratory} as an approach to compute numerical \emph{landscape features} for the purposes of problem characterization.  The original ELA features consisted of six classes; for instance, some of them quantify the distribution of the objective function values (\emph{y-Distribution}), while some others fit the meta-regression models (linear and quadratic) to the sampled data (\emph{Meta-Model}), etc.  ELA has since become an umbrella term for many newly introduced feature sets~\cite{lunacek2006dispersion, munoz2015ic, kerschke2015nbc}.  

Among different approaches to predict the algorithm performance on a certain problem as accurately as possible, supervised machine learning techniques, such as regression and classification, have been both well studied and used for PIAS in a variety of settings.  Performance \emph{regression} models predict the performances on a certain problem instance for the whole algorithm portfolio, and then proceed to select the algorithm with the best-predicted performance, thus keeping track of the magnitude of differences between different performances (something that would be lost if a classification approach was used).

\textbf{Our Results.} 
In recent years, most research efforts in automated algorithm selection were made within the \emph{fixed-target} context, where the algorithm performance is measured by the expected running time (that is, the number of function evaluations in the optimization context) needed to reach some pre-fixed solution of high quality, i.e., whose distance to the optimum is very small (e.g., in the order of $10^{-8}$).  Typically, a large budget of function evaluations (in the order of $10^4d$, where $d$ stands for dimension) is nevertheless allocated in the fixed-target setting to ensure that the run terminates. In real-world application and for practical purposes, however, it could be of vital importance to choose the algorithm that performs best within a given budget of function evaluations.  We adopt this \emph{fixed-budget} approach for our work, with a limited budget of only 500 function evaluations.  The quality of algorithm performance within the fixed-budget approach is measured by the \emph{target precision}, which is the distance between the optimal solution and the best found one.  

Our goal is to design a prediction model that learns the mapping between problem features on one hand and algorithm performance on the other hand, which will, once trained, be able to decide which algorithm to pick given a problem instance.  To the best of our knowledge, our work is the first to propose a novel performance regression model for single-objective, continuous problems in the fixed-budget setting, with a limited budget of function evaluations, which aims to solve the algorithm selection problem by combining the use of problem features that can be automatically computed and off-the-shelf supervised machine learning techniques.

In this work, we show that we are able to achieve high-quality performance prediction by combining two  differently trained regression models, one that predicts the target precision (``unscaled model'') and one that predicts the logarithm of the target precision (``logarithmic model'').  

\textbf{Related Work.} Automated algorithm selection and configuration can be roughly categorized into two classes, depending on whether they build upon supervised or unsupervised machine learning techniques. 
In terms of unsupervised learning, reinforcement learning in particular is used most frequently (see~\cite{biedenkapp2019,kerschke2018survey} and references mentioned therein).  
Among the supervised approaches, techniques that build on exploratory landscape analysis (ELA~\cite{mersmann2011exploratory}) predominate the field. ELA-based regression models have been applied both to \textit{configure} parametrized algorithm families (per instance algorithm configuration)~\cite{BelkhirDSS17} and to \textit{select} algorithms from a given portfolio~\cite{KerschkeT19,MunozKH12prediction}.

Comprehensive surveys of automated algorithm selection state-of-the-art approaches and results are available in~\cite{kerschke2018survey,munoz2015AS}.

\begin{figure*}
    \centering
    \includegraphics[width=\linewidth]{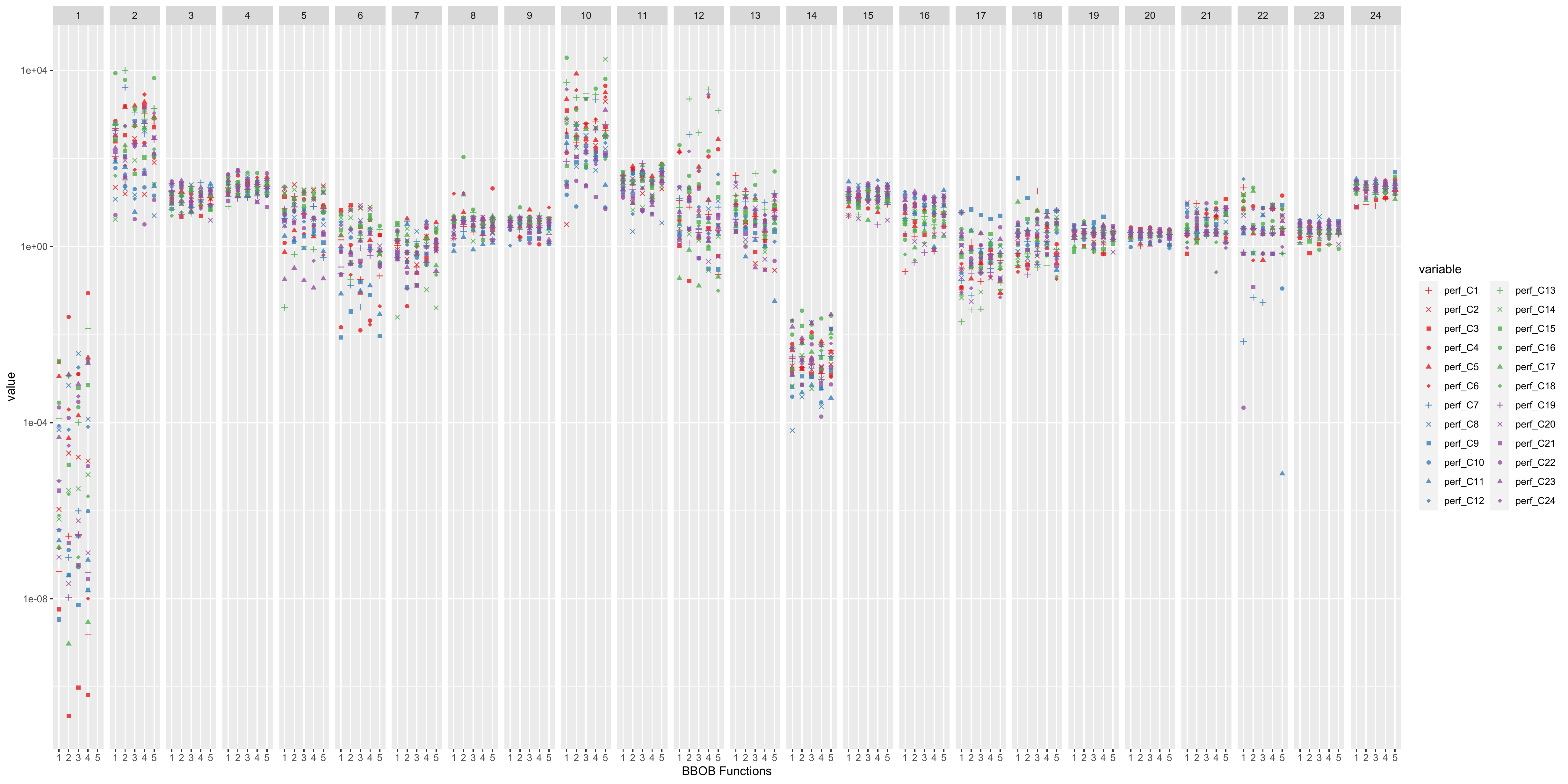}
    \caption{Median target precision of the 24 selected modular CMA-ES variants on the first four instances of all 24 BBOB functions. These 24 algorithms represent the algorithm portfolio from which we want to select the best-performing one for an unseen optimization problem.}
    \label{fig:performance}
\end{figure*}

\begin{figure}
    \centering
    \includegraphics[width=\linewidth]{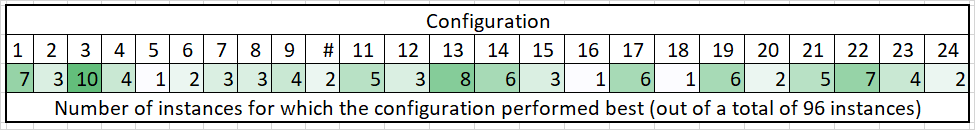}
    \caption{Number of problem instances (out of 96 total; second row) for which each of the algorithms (first row) achieved the best target precision.}
    \label{fig:winning-configs}
\end{figure}

\section{Experimental Setup}
\label{sec:setup}

\paragraph{The BBOB Testbed} 
Very well-known and widely used in the community, the \emph{Black-Box Optimization Benchmark} (BBOB) testbed is one of the benchmark problem suites on the \emph{COCO} platform~\cite{cocoplat}, which is the dedicated environment for comparison of algorithm performance in continuous black-box optimization.  
The BBOB testbed consists of 24 noiseless, single-objective functions (FID 1-24). For each function, different \emph{instances} can be generated by rotating or translating the function in the objective space.  All functions are defined and can be evaluated on $\mathbb{R}$, while the actual search space is given as $[-5, 5]^D$, where $D$ is the problem dimensionality.  We restrict our attention to the 5-dimensional variant of the first four instances (IID 1-4) of each function, which gives us a set of 96 different optimization problems in total.  An overview of the functions is available in~\cite{bbob-functions}.

\paragraph{The Modular CMA-ES} 
The Covariance Matrix Adaptation Evolution Strategy (CMA-ES~\cite{hansen2001self_adaptation_es}) is a popular and powerful Evolution Strategy heuristic, for which several modifications have been developed over the years in order to improve its performance for the large variety of different problems.  With motivation to present a unique framework able to generate different CMA-ES variants, a \emph{modular CMA-ES framework} was introduced in~\cite{van_rijn_evolving_2016} and later analyzed in~\cite{van_rijn_algorithm_2017}. It consists of 11 modules that a user can turn on or off as needed (some of the modules being, e.g., active update, elitism, constant population size/IPOP/BIPOP, etc). Nine of the modules are binary (can be turned on or off) and the remaining two, including the population size control, allow for three different choices. This leaves us with a total of 4608 modular CMA-ES variants. In the following, we interchangeably refer to these variants as \textit{variants}, \textit{configurations}, or simply \emph{algorithms}. The modular CMA-ES Python package is publicly available at~\cite{modCMAGitHubSander}.  

All 4608 modular CMA-ES variants were executed on the 96 problems mentioned above  (i.e., 4 instances per each of the 24 BBOB functions), with a budget of 500 function evaluations, for 5 independent replicated runs each.  After every run, we store the best target precision reached, computed as $f(x_{\text{best}}) - f(x_{\text{OPT}})$ (this value is positive, since we assume minimization as objective), as well as the function ID, instance ID, and algorithm ID.  We compute the median best target precision over different runs for all functions and instances and for each algorithm.  

From this median performance dataset, we select the best algorithms per function to create an algorithm portfolio that would act as the target data for our regression models.  To identify which algorithm is the best for a certain function, we compute the median performances of algorithms over \emph{instances} as well, which gives us 4608 different target precision values per function.  We then pick the algorithm with the minimum target precision among those 4608 for each function, and end up with a portfolio of 24 best algorithms in total.

It is important to recall that, unlike the fixed-target approach which typically makes use of the ERT (expected running time) as a performance metric, we operate within the fixed-budget approach, and consequently throughout this paper we use the \emph{target precision} after 500 function evaluations as a measure of an algorithm's performance.  Furthermore, the information carried by the target precision intuitively stands for the order of magnitude of the actual distance to the optimum.  For instance, if a recorded precision value on a certain problem instance is $10^{-2}$ for one algorithm and $10^{-8}$ for the other, and they differ by 6 orders of magnitude, then we can interpret that informally as the latter one being ``6 levels closer to the optimum'' than the first one.  This perspective helps immensely in designing the experiment, as we are interested not only in the actual precision values, but also in the ``distance levels'' to the optimum, which are very conveniently computed as the log-value of the precision.

We plot in Figure~\ref{fig:performance} the median target precision values (over 5 independent runs) of 24 different modular CMA-ES variants for each of the first four instances of the 24 BBOB functions.  Within this paper, this will  be our algorithm portfolio of choice.  We will use these values as the target data for our performance regression models. However, we observe that the 24 algorithms are in fact very similar in performance, which makes the algorithm selection problem in this setting rather challenging.  As shown in Figure~\ref{fig:winning-configs}, each algorithm ``wins'' at least one of the 96 considered instances.

\begin{figure*}[t]
    \centering
    \includegraphics[width=\linewidth]{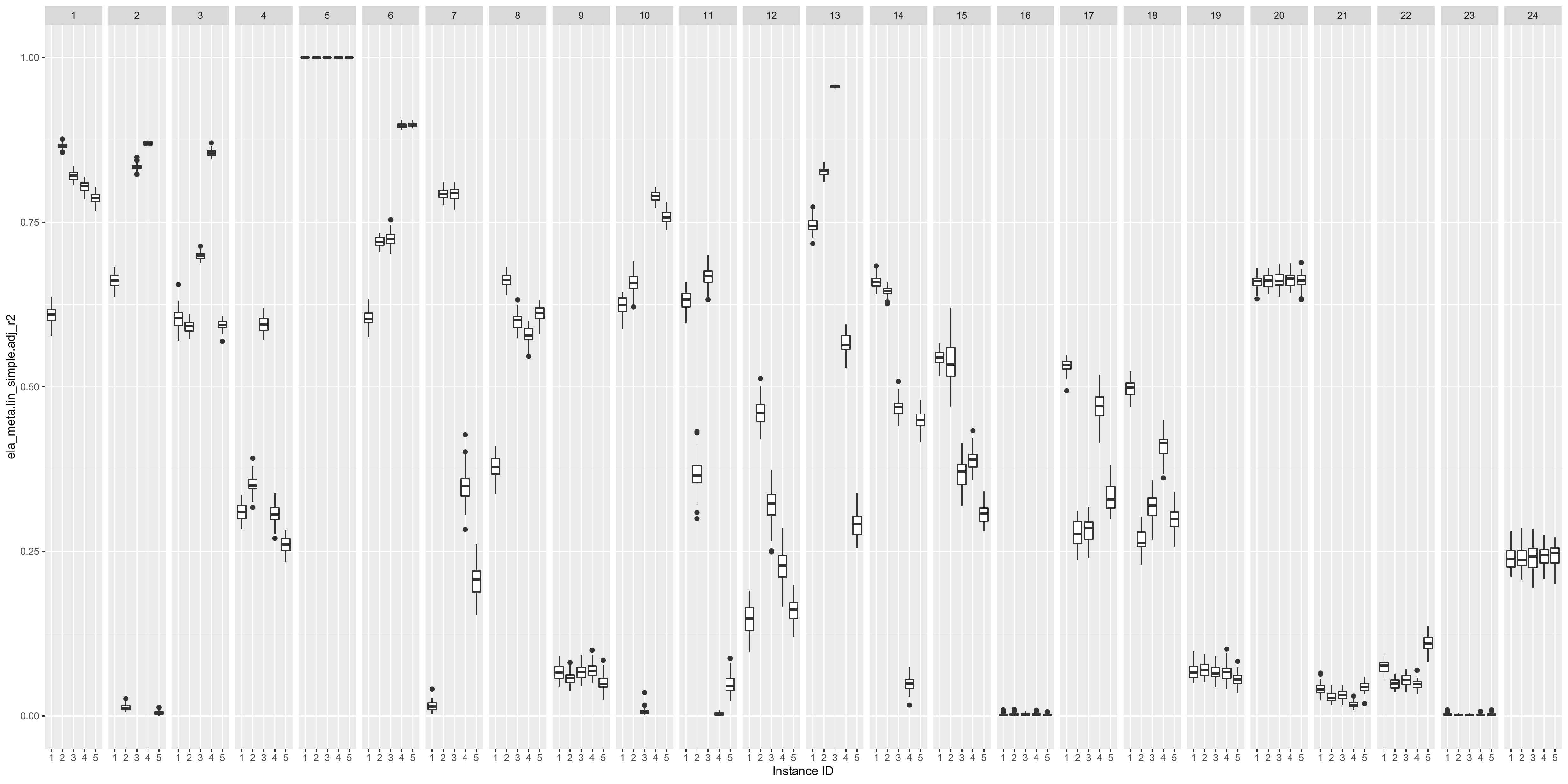}
    \caption{Distribution of approximated values for the \emph{ela\_meta\_lin\_simple\_adj\_r2} feature for the first five instances of the 24 BBOB functions, normalized in the 0-1 range. Values are computed from 2000 samples each, and each box plot shows the distribution of 50 independent runs.}
    \label{fig:features-ela_meta_lin_simple_adj_r2}
\end{figure*} 

\paragraph{Feature Computation} 
As the predictor variables for our model, we use vectors of landscape feature values per each problem instance
. For the feature value computation, we use the R package \emph{flacco}~\cite{flacco}, a publicly available toolbox that allows for the feature value approximations for all kinds of numerical optimization problems. 
To compute the features, we evaluate 2000 uniformly sampled search points per function and instance in 5D, and feed the points and their respective fitness values to \emph{flacco}. Since the feature approximations can show low robustness for certain features~\cite{GallagherECJ19TrueValue,Renau2019features}, we replicate this step 50 independent times and take the median feature values per problem instance as a final feature vector.  For the purpose of this work, we select only those feature sets that do not require further sampling in the search space: classical ELA ones (y-Distribution, Levelset and Meta-model)~\cite{mersmann2011exploratory}, Dispersion~\cite{lunacek2006dispersion}, Information Content~\cite{munoz2015ic} and Nearest-Better Clustering~\cite{kerschke2015nbc} feature sets.  That leaves us with 56 features in total per problem instance. 

It is worth noticing that the sample size of 2000 points might be much higher than one would be willing to invest in concrete applications, and we will therefore analyze the sensitivity of the results with respect to the sample size in Section~\ref{ssec:as-elasample}. Note here that -- in particular with such a small budget as investigated in our work -- the development of features that use the samples of a search trajectory rather than additional samples and hence avoid the specific sampling step for feature value computation~\cite{MunozS17footprint} is very strongly needed. 

We will also investigate the effect of the feature portfolio (cf. Section~\ref{ssec:as-thresh}), and show that a smaller set of features does not necessarily lead to worse results. 

Figure~\ref{fig:features-ela_meta_lin_simple_adj_r2} shows an example of how the feature values are distributed for the different instances. 
The feature in question is one of the classical ELA Meta-model features, a value of adjusted $R^2$ correlation coefficient for the linear model fitted to the data.  We see that the values are overall quite constant for most functions.  However, there are functions (e.g., F8, F12) for which this is not the case. All 5 instances of F5 are correctly identified as linear slope functions; they have a feature at value 1. Importantly, we observe that some features are more expressive than others, and are prone to discriminate between different problems better than others, as suggested by~\cite{Renau2019features}.

\paragraph{Random Forest Model} 
For our regression models, we use an off-the-shelf random forest regressor with 1000 estimators using the  \emph{scikit-learn} Python package~\cite{scikit-learn}. A random forest is an ensemble-based meta-estimator that fits decision trees on various sub-samples of the data set and uses averaging to improve the predictive accuracy and to control over-fitting.  No parameter tuning was involved in our setup, which gives us hope that a further improvement of the accuracy is highly likely by fine-tuning the machine learning model, but also that a clever design of underlying model mechanism can render the processing itself quite cheap.

\section{Fixed-Budget Performance Regression}
\label{sec:regression}

Using the elements described in Section~\ref{sec:setup} as key pieces in our approach, we organize the experiment in the following way.  For each algorithm in the portfolio, we train two separate random forest models to predict its performance on all problem instances, given a vector of features per problem instance.  One model is trained on the actual precision values as target data (we call it the \emph{unscaled model} onward), while the other uses log-target precision data to represent the ``distance levels'' explained in Section~\ref{sec:setup} (we refer to this model as the \emph{logarithmic model} or simply the \emph{log model}).  

In order to obtain more realistic and reliable estimates of the models' accuracy, we assess them using a $K$-fold \emph{leave-one-instance-out} cross-validation (we use $K = 4$), i.e., per algorithm, we split the data so that we use three instances per BBOB function for the training (72 problems in total) and we test on the remaining instance (24 problems in total).  We do this with each of the four instances to ensure that each instance was used once in the test phase.  We consistently store the values of the prediction on the test instance for each algorithm in the portfolio for both the unscaled and the log-model. The full experiment is replicated three independent times and the median values are taken for the analysis.

\begin{figure*}
    \centering
    \includegraphics[width=\linewidth]{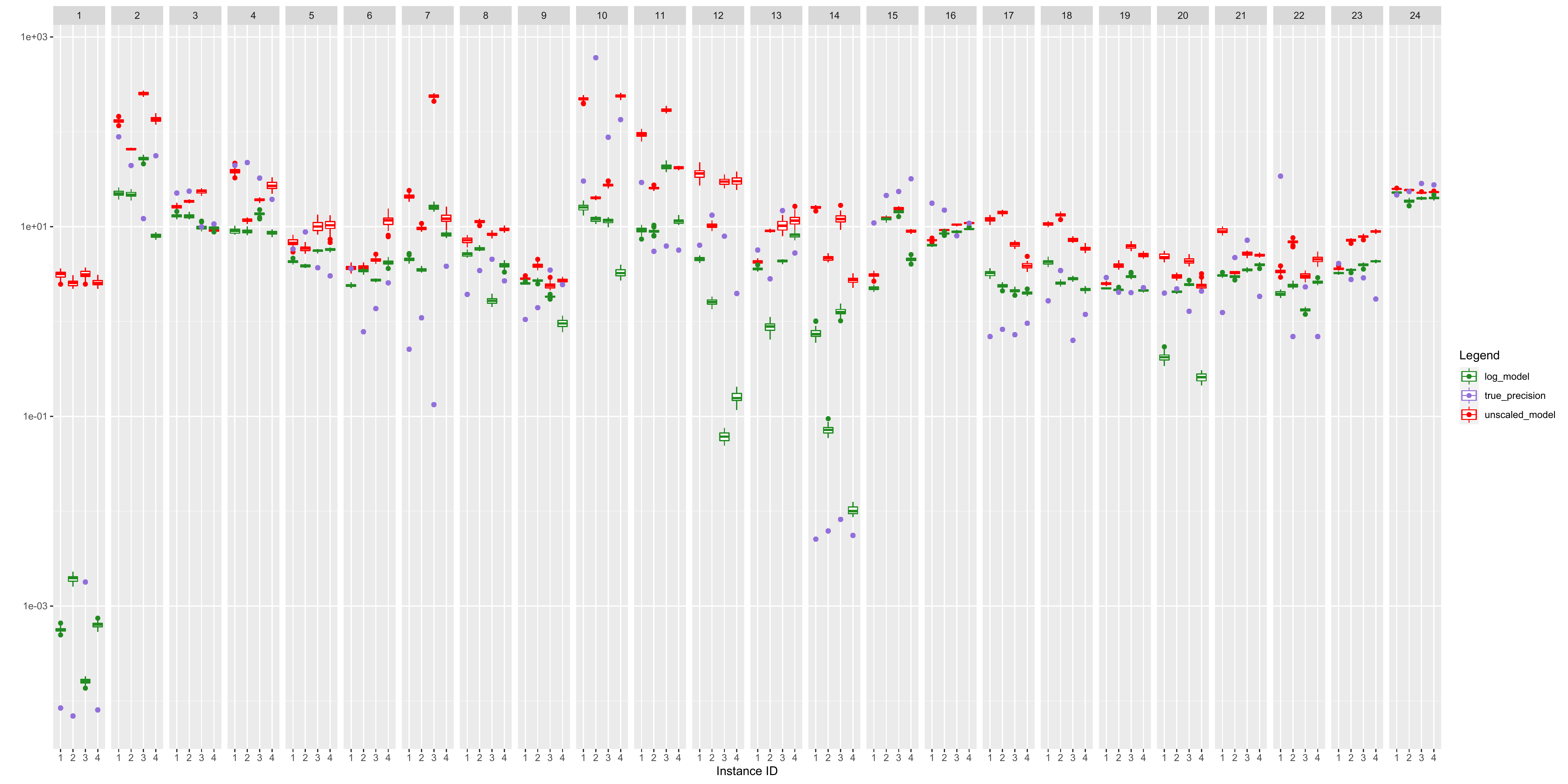}
    \caption{Distribution of predicted values of the unscaled (in red) and logarithmic regression models (in green) across 50 independent runs, with the actual target precision values (in purple) as dots.}
    \label{fig:config12}
\end{figure*}

We plot in Figure~\ref{fig:config12} true vs. predicted values of a single variant of the modular CMA-ES for 50 independent runs for each of the first four instances of the 24 BBOB functions.  The distribution of two different predictions, unscaled and log-prediction, are shown.  We observe a  high stability of the predictions irrespective of the number of replications (i.e., whether we do 3 runs, as throughout our experiments, or 50 runs, as shown here), and thus conclude that this allows for a significant decrease in the computational cost of creating such a model. 

Importantly, Figure~\ref{fig:config12} shows how well the predicted values follow the actual data.  As a general trend, the predictions of the unscaled model fit better to larger precision values, while the log-model better predicts small target precisions.

\begin{table}[h]
\resizebox{\columnwidth}{!}{%
\begin{tabular}{@{}c|cc|cc|cc@{}}
\toprule
        & \multicolumn{2}{c|}{Default}                & \multicolumn{2}{c|}{Selected features}  & \multicolumn{2}{c}{50$d$
        samples} \\ \midrule
Config. & RMSE                 & log-RMSE             & RMSE            & log-RMSE       & RMSE             & log-RMSE     \\ \midrule
C1      & \textbf{130.1}       & {\ul 0.808}          & 135.4           & \textbf{0.663} & {\ul 123.6}      & 1.066        \\
C2      & 84.1                 & {\ul 0.776}          & \textbf{79}     & \textbf{0.682} & {\ul 77.7}       & 0.891        \\
C3      & 206.5                & {\ul 0.842}          & \textbf{199.7}  & \textbf{0.663} & {\ul 189.6}      & 1.308        \\
C4      & 201.9                & {\ul 0.757}          & \textbf{198.3}  & \textbf{0.682} & {\ul 200.7}      & 0.829        \\
C5      & 1011.2               & {\ul 0.690}          & \textbf{916.8}  & \textbf{0.564} & {\ul 1106.6}     & 0.742        \\
C6      & \textbf{649.9}       & {\ul \textbf{0.986}} & 660.6           & 1.018          & {\ul 618.1}      & 1.030        \\
C7      & 462.8                & {\ul 0.804}          & \textbf{455}    & \textbf{0.743} & {\ul 412.8}      & 0.989        \\
C8      & 61.2                 & {\ul 0.830}          & \textbf{58.8}   & \textbf{0.698} & {\ul 57.4}       & 0.925        \\
C9      & {\ul \textbf{71.1}}  & {\ul 0.770}          & 74.3            & \textbf{0.623} & 77.3             & 1.133        \\
C10     & {\ul 12.1}           & {\ul 0.771}          & \textbf{11.4}   & \textbf{0.649} & 12.2             & 1.080        \\
C11     & {\ul 71.9}           & {\ul 0.794}          & \textbf{55.2}   & \textbf{0.653} & 78.3             & 1.060        \\
C12     & {\ul 76.7}           & {\ul 0.708}          & \textbf{64.9}   & \textbf{0.601} & 78.1             & 0.789        \\
C13     & \textbf{1120.4}      & {\ul 0.700}          & 1124.7          & \textbf{0.696} & {\ul 1113.1}     & 0.789        \\
C14     & \textbf{51.1}        & {\ul 0.792}          & 51.4            & \textbf{0.678} & {\ul 44.4}       & 0.973        \\
C15     & 60.5                 & {\ul 0.629}          & \textbf{54.7}   & \textbf{0.519} & {\ul 56.5}       & 0.748        \\
C16     & 2306.3               & {\ul 0.621}          & \textbf{2280.6} & \textbf{0.604} & {\ul 2239.0}     & 0.791        \\
C17     & 114.3                & {\ul 0.781}          & \textbf{98}     & \textbf{0.631} & {\ul 111.2}      & 1.134        \\
C18     & {\ul \textbf{130.4}} & {\ul 0.640}          & 149.6           & \textbf{0.596} & 131.6            & 0.903        \\
C19     & 85.1                 & {\ul 0.710}          & \textbf{82.4}   & \textbf{0.571} & {\ul 73.5}       & 1.025        \\
C20     & \textbf{144.3}       & {\ul 0.760}          & 152.9           & \textbf{0.618} & {\ul 138.7}      & 1.032        \\
C21     & 23.2                 & {\ul 0.719}          & \textbf{23}     & \textbf{0.662} & {\ul 20.7}       & 1.007        \\
C22     & 17.0                 & {\ul 0.805}          & \textbf{16.4}   & \textbf{0.714} & {\ul 16.5}       & 0.919        \\
C23     & {\ul 53.6}           & {\ul 0.613}          & \textbf{45.8}   & \textbf{0.538} & 55.0             & 0.691        \\
C24     & \textbf{571.9}       & {\ul \textbf{0.803}} & 604.7           & 0.86           & {\ul 531.1}      & 0.872        \\ \bottomrule
\end{tabular}%
}
\caption{Root Mean Square Error (RMSE) for the unscaled and log-model as a measure for model prediction accuracy for each algorithm in the portfolio in 3 different scenarios.  They compare how well different models fit the actual target data.  The default experiment (the first 2 columns) consists of the performance regression using the full feature set, where features were computed using 2000 samples.  The second 2 columns correspond to the experiment where the regression was based on \emph{9 selected features only}, while the third 2 columns describe the case where, again, the full feature set was used, but this time the features were \emph{computed using 50$d$ (250) samples}. The values shown in bold represent lower errors when comparing the first 2 scenarios (all features and selected features), while the underlined values highlight lower errors when the 1$^{st}$ and the 3$^{rd}$ scenario are compared (2000 samples and 250 samples).}
\label{tab:rmseperconfig-anja}
\end{table}

As a measure of model accuracy, the \emph{Root Mean Square Errors} (RMSE) and log-RMSE are computed per prediction.  They allow us to compare (and quantify) how well different models predict the performance.  The RMSE metric is the standard deviation of prediction errors; it measures how spread out those errors are, and is frequently used.  In regression, the prediction errors are the distance of the prediction to the true target value.  Within the regression context, while the RMSE corresponds to the unscaled model, the log-RMSE corresponds to the log-model.  Once computed, we aggregate the relevant RMSE and log-RMSE per algorithm to estimate the quality of the model at hand.  

We report in the first two columns of Table~\ref{tab:rmseperconfig-anja} how good the random forest models (unscaled and log) were at predicting each algorithm's performance. We observe that the same algorithm (e.g., C16) can have high RMSE (and thus be very bad in predicting the actual data) all while having one of the lowest log-RMSE (quite the opposite for the log-data).  

\subsection{Impact of Feature Selection}
\label{ssec:regr-sens}

We have reported above results for the regression models that make use of 56 \emph{flacco} features. However, it is well known that feature selection can significantly improve the accuracy of random forest regressions. We therefore present a brief sensitivity analysis, which confirms that significantly better results can be expected by an appropriate choice of the feature portfolio. Note that a reduced feature portfolio also has the advantage of faster computation times, both in the feature extraction and in the regression steps. We have performed an adhoc feature selection, which solely builds on a visual analysis of the approximated feature values. This was done by studying the plots as in Figure~\ref{fig:features-ela_meta_lin_simple_adj_r2} for all computed \emph{flacco} features, which we use to identify features that show a high \emph{expressiveness}~\cite{Renau2019features}, in the sense that they seem very suitable to discriminate between the different BBOB functions. From this set, we then choose nine features: \emph{disp.diff\_mean\_02, ela\_distr.skewness, ela\_meta.lin\_simple.adj\_r2, ela\_meta.lin\_simple.coef.max, ela\_meta.lin\_simple.intercept, ela\_meta.quad\_simple.adj\_r2, ic.eps.ratio, ic.eps.s} and \emph{nbc.nb\_fitness.cor}. 

The middle two columns of Table~\ref{tab:rmseperconfig-anja} illustrate how the model accuracy indeed increases with the reduced feature set. 16 out of the 24 RMSE values and 22 of the 24 log-RMSE values are smaller than for the model using all features (which we recall are provided in the 2 leftmost columns of the same table).  The values of the more accurate model between the two are highlighted in bold in this case. These results support the idea that an appropriate feature selection is likely to result in significant improvements of our regressions, and hence of the algorithm selectors which we shall discuss in the next section.

\section{Fixed-Budget Algorithm Selection}
\label{sec:selection}

After examining the regression models' accuracy, we next evaluate the performance of two simple algorithm selectors, which are based on the predictions of the unscaled and the logarithmic models, respectively.  The former selects the algorithm for which the unscaled regression model predicted the best performance, and, similarly, the latter bases its decision on the best prediction of the log-regression model.  To quantify how well the selectors perform per problem instance, we compare the precision of the algorithm chosen by the selector for the instance at hand to the precision of the actual best algorithm for that instance.  We are then able to indicate the overall quality of this selector by computing the RMSE and log-RMSE values (aggregated across all problems) using the differences in performance.  

Following common practices in algorithm selection~\cite{bischl_aslib:_2016}, we compare the performances of these two selectors with two different baselines. The \emph{virtual best solver} (VBS, also called the oracle) gives a lower bound for the selectors. It always chooses the best-performing algorithm per each problem instance without any additional information or cost, hence it reflects the best performance that could be theoretically achieved. 

On the other hand, the \emph{single best solver} (SBS) represents the overall (aggregated) performance of the best-performing algorithm from the portfolio. Since we have two models, we have two different SBS, one for the log-model (SBS$_{\log}$) and one for the unscaled one (SBS$_{\text{unscaled}}$). Studying the RMSEs of the 24 different algorithms (Figure~\ref{fig:sbstable}), we find that configuration C10 has the best performance (measured against the VBS), with an RMSE value of 13.65.   
Configuration C21, in contrast, is seen to have the best log-RMSE value, and is therefore the SBS$_{\log}$ of the full portfolio. Its log-RMSE value is 0.733. In the following, for ease of notation, we will not distinguish between the two SBS, and, in abuse of notation, will combine them into one. That is, we simply speak of the SBS, and refer to C10 when discussing RMSE values, while we refer to C21 when discussing log-RMSE values.   
As we can see in Figure~\ref{fig:sbstable}, our algorithm selectors are able to outperform the SBS in terms of log-RMSE performance. We did not find a way, however, to beat the RMSE-values of C10, nor the ones of C8, C14, C21, C22, nor C23.   

Performances of single algorithms from the portfolio, as well as those of the two selectors (unscaled and log-selector) are shown in Figure~\ref{fig:sbstable}. For the majority of the portfolio, a lower RMSE value entails a lower log-RMSE value and vice versa.  We see that our two selectors already outperform the majority of algorithms from the portfolio on both RMSE- and log-RMSE scales, but we want to make use of the observation reported in Section~\ref{sec:regression} that the unscaled model better predicted higher target precision, while the log-model has better accuracy for small target precision. 
We therefore aim at combining the two regression models, to benefit from the two complementing strengths. Here again we can define a virtual best solver, which is the one that chooses for each instance the better of the two suggested algorithms from the unscaled and the logarithmic model, respectively. Clearly, in terms of single best solver, the log-model minimized log-RMSE, whereas the unscaled model minimizes RMSE. 
We compare these three algorithm selectors (unscaled AS, logarithmic AS, and VBS AS) with a selector which combined the two basic selectors in the following way: if the target precision of an algorithm, as predicted by the log-model, is smaller than a certain threshold, we use the log-selector, whereas we use the recommendation of the unscaled AS otherwise. A new optimization sub-problem that arises here is to find the threshold value which minimizes RMSE and log-RMSE of the combined selector, respectively. A sensitivity analysis with respect to this threshold will be presented in Section~\ref{ssec:as-thresh}. 

Once the optimal threshold value found, we measure the performance of our selector and add it, along with the VBS AS, to Figure~\ref{fig:sbstable}.  We clearly see that our algorithm selector performs better than the unscaled AS and the logarithmic AS. It is also better than most of the algorithms.  
Also, Figure~\ref{fig:sbstable} demonstrates that our selector effectively reduces the gap towards the VBS AS. Detailed numbers for the RMSE and log-RMSE values of the different algorithm selectors are provided in Table~\ref{tab:sens-portfolio} (rows for 24 algorithms). 

\begin{figure*}
    \centering
    \includegraphics[width=\linewidth]{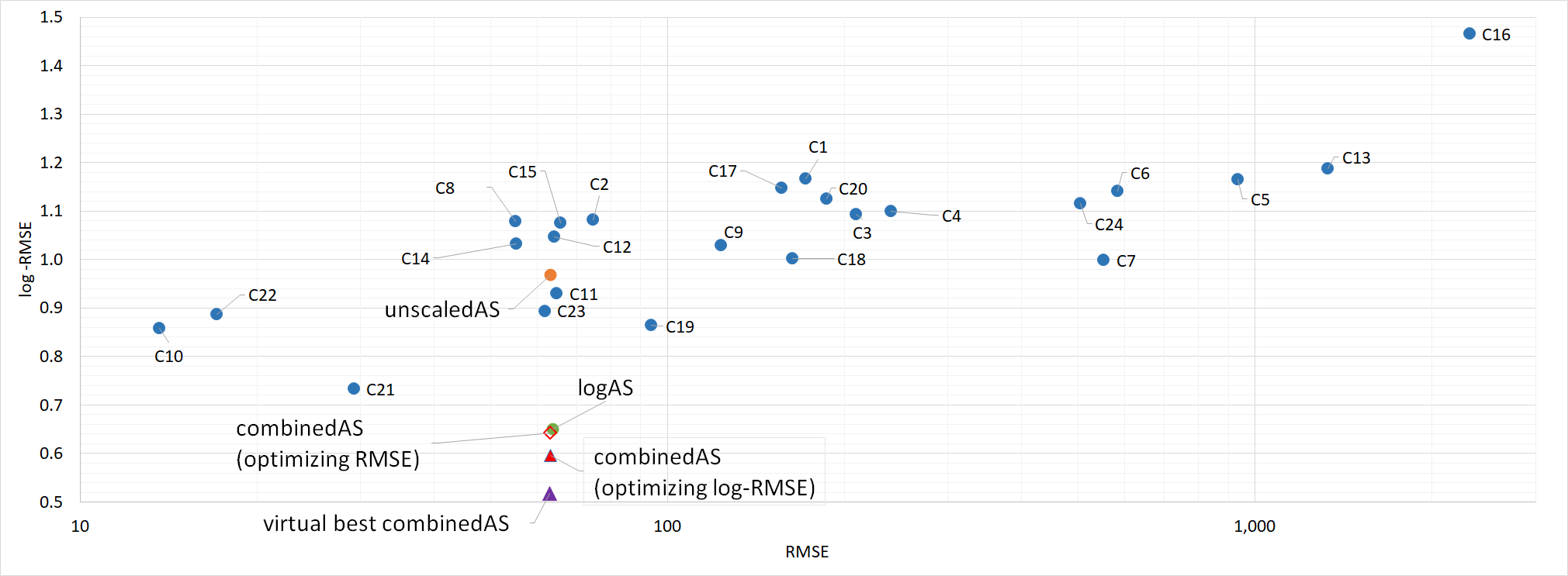}
    \caption{RMSE and log-RMSE values as measures for the quality of the configurations (label C$x$) and for the algorithm selectors: the one using only the predictions from the log model (green dot, hiding behind red diamond), the one using predictions from the unscaled model (orange dot), the virtual best combination of these two models (purple triangle), and our two combined selectors, which optimize for RMSE and log-RMSE (red diamond and red triangle, respectively). }
    \label{fig:sbstable}
\end{figure*}

\subsection{Impact of the Threshold Value and the Feature Portfolio}
\label{ssec:as-thresh}

We now study the influence of the threshold value which determines whether we use the unscaled of the log-model algorithm selector. In the previous section, we had chosen this value so that it optimized the log-RMSE measure and the RMSE-measure (these are the red triangle and the red diamond in Figure~\ref{fig:sbstable}, respectively). 
Table~\ref{tab:sens-thresh} analyzes the influence of this threshold value, and shows both RMSE and log-RMSE values for different thresholds. We note that one could formulate an alternative decision rule, in which the selection of the model is not based on the target precision recommended by the logarithmic model, but by the unscaled model. We did not observe significant differences in the performance of these two approaches and omit a detailed discussion for reasons of space. 

We show in this table also the results for the algorithm selectors that build on the regression models using only the nine selected features. In fact, it turns out that for these selected-feature regression models, the combination of the two different regressions is not beneficial -- we were not able to identify means to improve upon the algorithm selector that uses the log-model-predicted fixed-budget performances. 

\begin{table}[h]
\centering
\resizebox{\columnwidth}{!}{%
\begin{tabular}{@{}c|cc|cc@{}}
\toprule
          & \multicolumn{2}{c|}{\textbf{RMSE}}        & \multicolumn{2}{c|}{\textbf{log-RMSE}}     \\ \midrule
\textbf{Threshold} & \textbf{All features} & \textbf{Selected features} & \textbf{All features} & \textbf{Selected features} \\
\midrule
0.01      & 63.20        & 25.87             & 0.687        & 0.728             \\
0.1       & 63.20        & 25.87             & 0.676        & 0.723             \\
0.5       & 63.19        & 25.81             & 0.600        & 0.637             \\
0.814     & 63.19        & 25.81             & \underline{\textbf{0.595}}        & 0.627             \\
1         & 63.19        & 25.77             & 0.624        & 0.620             \\
2         & 63.17        & 25.81             & 0.643        & 0.607             \\
2.294         & \underline{\textbf{63.17}}      &     25.80      &    0.643     &         0.590     \\
3         & 63.75        & 25.82             & 0.656        & 0.583             \\
8.525     & 63.71             & \underline{\textbf{15.50}}       & 0.654          & \underline{\textbf{0.565}} \\
10        & 63.71        & 15.55             & 0.654        & 0.565             \\
20        & 63.69        & 15.55             & 0.650        & 0.565             \\
50        & 63.69        & 15.55             & 0.650        & 0.565             \\ \bottomrule
\end{tabular}
}
\caption{Sensitivity of the RMSE and the log-RMSE with respect to the threshold value at which we switch from choosing the log-model-suggested CMA-ES configuration to the one suggested by the unscaled model, for both the models using the full feature set for regression and the one using the selected features only. Optimal values are shown in bold and are underlined.}
\label{tab:sens-thresh}
\end{table}

\subsection{Impact of the Algorithm Portfolio}
\label{ssec:as-portfolio}

Our final analysis concerns the  portfolio for which we do the regression. Note that in all the above we have given ourselves a very difficult task: algorithm selection for a portfolio of 24 solvers that all show quite similar performance (Figure~\ref{fig:performance}). We now study alternative problems, in which we consider only subsets of the 24 CMA-ES configurations considered above. More precisely, we consider in Table~\ref{tab:sens-portfolio} the portfolio of configurations C13-C24, i.e., the second half of the original portfolio. 

We observe that, while the log-RMSE values remain fairly consistent for all the selectors independently of the portfolio size, the RMSE is significantly reduced by reducing the portfolio size for all but one selector, the SBS. It is worth looking further into this aspect of the problem in order to better understand if the observed effects are simply a product of less choice, or, more likely, there are other factors at play, e.g., whether the algorithms chosen for the portfolio are diverse enough in their performance on different problem instances.

\begin{table}[h]
\resizebox{\columnwidth}{!}{%
\begin{tabular}{@{}c|c|ccccc@{}}
\toprule
                          & \# algos & unscaled & log   & the AS & VBS   & SBS   \\ \midrule
\multirow{2}{*}{RMSE}      
                            & 12     & 17.25    & 18.03 & 16.94  & 12.78 & 20.37  \\
                            & 24     & 63.19    & 63.69 & 63.19  & 63.09 & 13.65  \\ \midrule
\multirow{2}{*}{log-RMSE}   
                            & 12     & 0.967    & 0.621 & 0.608  & 0.561 & 0.629  \\
                            & 24     & 0.968    & 0.650 & 0.595  & 0.517 & 0.733  \\ \bottomrule
\end{tabular}
}
\caption{Comparison of the RMSE and log-RMSE values for the different algorithm selectors for the full portfolio of 24 and for the reduced set of 12 configurations.}
\label{tab:sens-portfolio}
\end{table}

\subsection{Impact of the Feature Sample Size}
\label{ssec:as-elasample}

 We recall that our feature approximations are based on 2000 samples. As commented above, this number is much larger than what one could afford in practice. Belkhir et al.~\cite{BelkhirDSS17} showed that sample sizes as small as 30$d$-50$d$ can suffice to obtain reasonable results. While their application is in algorithm \emph{configuration}, we are interested in knowing whether we obtain similarly robust performance for algorithm \emph{selection}. As mentioned before, in the long run, one might hope for zero- or low-cost feature extraction mechanisms that simply use the search trajectory samples of a CMA-ES variant (or some other solver) to predict algorithm performances and/or perform a selection task. First steps in this direction have already been made~\cite{MunozS17footprint,MalanM19,Malan18}. 
 
 Therefore, in this last section we study the influence of the feature sample size on the performance of our algorithm selector. We compare the results reported in Section~\ref{sec:selection} with the results obtained from the repeated regression experiment, only this time using features computed with 50$d$ (250) samples.
 
\begin{table}[h]
\resizebox{\columnwidth}{!}{%
\begin{tabular}{@{}c|c|ccccc@{}}
\toprule
                          & \# feature samples & unscaled & log   & the AS & VBS   & SBS   \\ \midrule
\multirow{2}{*}{RMSE}      
                            & 250     &  23.51    &  38.74 & 23.45  & 23.05 & 13.65  \\
                            & 2000     & 63.19    & 63.69 & 63.19  & 63.09 & 13.65  \\ \midrule
\multirow{2}{*}{log-RMSE}   
                            & 250     & 0.881    & 0.700 & 0.660  & 0.511 & 0.733  \\
                            & 2000     & 0.968    & 0.650 & 0.595  & 0.517 & 0.733  \\ \bottomrule
\end{tabular}
}
\caption{Comparison of the RMSE and log-RMSE values for the different algorithm selectors for the regression based on 250-sample features and 2000-sample features.}
\label{tab:sens-samplesize}
\end{table}
 
The two rightmost columns of Table~\ref{tab:rmseperconfig-anja} show the regression model accuracy for this case, and allow for a comparison with the default scenario.  We once again conveniently highlight the values of the more accurate model, only this time underlined.  We clearly notice that, in terms of the RMSE values of the log-model, using a larger sample size is preferable consistently for all the algorithms in the portfolio.  On the other hand, the model using a reduced sample size performed better on 18 out of 24 algorithms in terms of RMSE. 
 
In Table~\ref{tab:sens-samplesize} we report the differences of the algorithm selectors in case the regression was based on features computed using 50$d$ (250) samples vs. those computed using the original 2000 samples.  The results are comparable in terms of both RMSE and log-RMSE values; the distances between the performance of our combined selector and the VBS are similar in both 250- and 2000-sample experiments. Also, in neither of the two experiments have we been able to beat the SBS.  However, we notice a general decrease in the RMSE when using 250 samples to compute the features, which is an interesting observation leading to a conjecture that it might be preferable to use a smaller number of samples to compute the features for regression purposes, while still maintaining the robustness of the results.  

\section{Conclusions}
\label{sec:conclusions}

We have studied in this work how to increase the accuracy of ELA-based regression models and algorithm selection by combining a plain ``unscaled'' regression with a regression operating on the log-scaled data.  While the former achieves higher accuracy for large target precision values, the latter performs better for fine-grained precisions.  By combining the two models, we could improve the accuracy of the regression and of the algorithm selector.  Our combined AS reduces the gap towards the VBS, although it does not consistently beat the SBS across all cases.  These results, however, still open up a path to further exploit the power of ELA-based regression and algorithm selection in different settings.

In the remainder of this paper, we list a few promising avenues for future work. 

\emph{Cross-Validation of the Trained Algorithm Selector on Other Black-Box Optimization Problems.} Our ultimate goal is to train an algorithm selector that performs well on previously unseen problems. We are therefore keen on testing our regression models for the different CMA-ES variants and on testing the trained algorithm selector on other benchmark functions. The current literature is not unanimous w.r.t. to the quality that one can expect from the training on the BBOB functions. While~\cite{BelkhirDSS17} reported encouraging performance, LaCroix and McCall~\cite{LacroixM19} could not achieve satisfactory results.  

\emph{Feature Selection.} The results presented in Section~\ref{ssec:regr-sens} indicate that a proper selection of the features can improve the quality of the random forest regression quite significantly. Our feature selection was based on an purely visual interpretation of the distribution of the feature value approximations (i.e., plots as in Fig.~\ref{fig:features-ela_meta_lin_simple_adj_r2}), which is similar to the analyses made in~\cite{Renau2019features,MunozReliability}. A proper feature selection may help to improve the accuracy of our models further. Since feature selection is quite expensive in terms of computational cost, a first step could be a comparison of the accuracy of the two here-presented models with those using the feature sets selected in~\cite{KerschkeT19}. 

\emph{Fixed-Target Settings.} While we have deliberately chosen a fixed-budget setting (which is the setting of our envisaged applications), we are nevertheless confident that the combination of a logarithmic with an unscaled regression model could also prove advantageous in fixed-target settings, in which the goal is to minimize the average time needed to identify a solution of function value at least as good as some user-defined threshold.

\emph{Different Algorithm Portfolios.} We have chosen a very challenging task in performing algorithm selection on a collection of algorithms that all stem from the same family. A cross-validation of our findings on more diverse portfolios is a straightforward next step for our work.

\begin{acks}
We thank Diederick Vermetten for sharing the performance data of the modular CMA-ES and Quentin Renau for suggesting the limited-size feature portfolio. We also thank Pascal Kerschke for help with the flacco package. 

Our work was financially supported by the Paris Ile-de-France Region and by a public grant as part of the
Investissement d'avenir project, reference ANR-11-LABX-0056-LMH,
LabEx LMH.  

We also acknowledge support from COST action CA15140 on Improving Applicability of Nature-Inspired Optimisation by Joining Theory and Practice (ImAppNIO). 
\end{acks}

} 


\end{document}